\pdfoutput=1

\documentclass[11pt]{article}

\usepackage[final]{acl}

\usepackage{times}
\usepackage{latexsym} 

\usepackage[T1]{fontenc}

\usepackage[utf8]{inputenc}

\usepackage{microtype}

\usepackage{inconsolata}

\usepackage{graphicx}

\usepackage{xcolor}
\usepackage{times}
\usepackage{soul}
\usepackage{url}
\usepackage{caption}
\usepackage{amsmath}
\usepackage{todonotes}
\usepackage{amsthm}
\usepackage{booktabs}
\usepackage{algorithm}
\usepackage{algorithmic}
\usepackage{pdfpages}
\usepackage{multirow}
\usepackage{amsthm}
\usepackage{makecell}

\usepackage[switch]{lineno}
\usepackage{xurl}

\title{Recent Advances in Online Hate Speech Moderation:\\Multimodality and the Role of Large Models}

\author{
 \textbf{Ming Shan Hee\textsuperscript{1}\thanks{These authors contributed equally to this work.}},
 \textbf{Shivam Sharma\textsuperscript{2,*}},
 \textbf{Rui Cao\textsuperscript{3}},
 \textbf{Palash Nandi\textsuperscript{2}}, \\
 \textbf{Preslav Nakov\textsuperscript{4}},
 \textbf{Tanmoy Chakraborty\textsuperscript{2}},
 \textbf{Roy Ka-Wei Lee\textsuperscript{1}},
\\
 \textsuperscript{1}SUTD,
 \textsuperscript{2}IIT Delhi,
 \textsuperscript{3}SMU,
 \textsuperscript{4}MBZUAI
\\
 \textsuperscript{1}\small{\{mingshan\_hee@mymail., roy\_lee@\}sutd.edu.sg} \\
 \textsuperscript{2}\small{\{shivam.sharma, palash.nandi, tanchak\}@ee.iitd.ac.in} \\ 
 \textsuperscript{3}\small{ruicao.2020@phdcs.smu.edu.sg} \\
 \textsuperscript{4}\small{preslav.nakov@mbzuai.ac.ae}
 }

\begin{document}
\maketitle

\begin{abstract}
Moderating hate speech (HS) in the evolving online landscape is a complex challenge, compounded by the multimodal nature of digital content. This survey examines recent advancements in HS moderation, focusing on the burgeoning role of large language models (LLMs) and large multimodal models (LMMs) in detecting, explaining, debiasing, and countering HS. We begin with a comprehensive analysis of current literature, uncovering how text, images, and audio interact to spread HS. The combination of these modalities adds complexity and subtlety to HS dissemination. We also identified research gaps, particularly in underrepresented languages and cultures, and highlight the need for solutions in low-resource settings. The survey concludes with future research directions, including novel AI methodologies, ethical AI governance, and the development of context-aware systems. This overview aims to inspire further research and foster collaboration towards responsible and human-centric approaches to HS moderation in the digital age.\footnote{ \textcolor{red}{WARNING: This paper contains offensive examples.}} 

\end{abstract}

\begin{figure}[t!]
\centering
\includegraphics[width=1.0\linewidth]{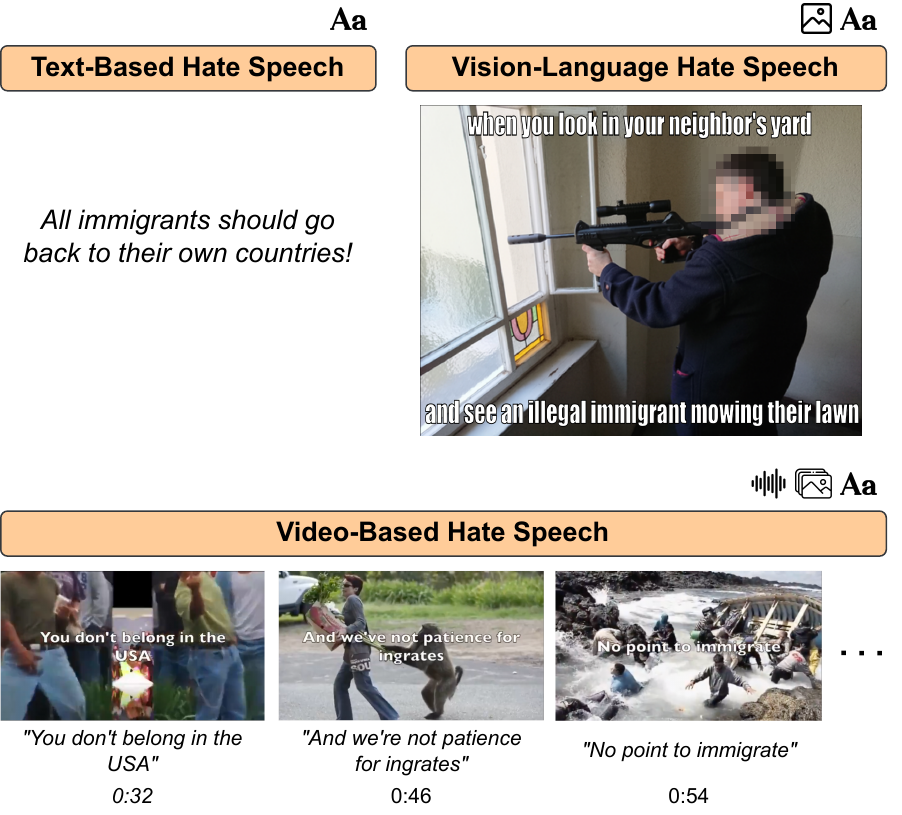}
\caption{Examples of an anti-migrant HS in different forms, encompassing text, image and/or audio modalities. The text-based, vision-language and video-based HS are taken from the Social Bias Inference Corpus (SBIC) dataset, the Facebook Hateful Memes (FHM) dataset and the Bitchute website, respectively.}
\label{fig:hate-speech-modalities}
\end{figure}

\section{Introduction}
In the era of rapid information exchange and digital connectivity, the rise of hate speech (HS) presents a significant challenge with profound implications for global societies. HS, which is any communication demeaning a person or a group based on social or ethnic characteristics, undermines social harmony and individual safety, both online and offline \cite{lupu2023offline}. The recent Israel--Hamas conflict has notably escalated both anti-Muslim and anti-Semitic sentiments
worldwide, evidenced by the trending of hashtags such as \textit{\#HitlerWasRight} and \textit{\#DeathToMuslim} on the social media platform X.\footnote{\url{https://www.nytimes.com/2023/11/15/technology/hate-speech-israel-gaza-internet.html}} Moreover, the Council on American--Islamic Relations reported receiving 774 help requests and bias reports from Muslims in the USA within a 16-day period.\footnote{\url{https://www.cair.com/press\_releases/cair-reports-sharp-increase-in-complaints-reported-bias-incidents-since-107/}} While digital interconnectivity facilitates swift information sharing, it simultaneously amplifies the spread and the impact of HS, transcending geographical boundaries.

Technological advancements have transformed the expression of HS, leading to its manifestation in various novel forms. Traditionally, HS was predominantly text-based, found in written materials \cite{rini2020systematic}, or verbalized in posts, broadcasts, and public speeches \cite{nielsen2002subtle}. The digital era has ushered in more complex and subtle variants of HS, engaging multiple sensory modalities. A notable instance is vision-language HS, which fuses visual elements with text, commonly disseminated through captioned images and memes \cite{uyheng2020visualizing, DBLP:conf/nips/KielaFMGSRT20}. Video-based HS, another emerging form, amalgamates text, visuals, and audio, creating a multi-faceted and potentially more influential mode of communication \cite{das2023hatemm}. Figure~\ref{fig:hate-speech-modalities} shows various HS forms targeting immigrants, underscoring animosity towards individuals of diverse nationalities. The text-based approach overtly projects hostile attitudes towards them in the host country. In vision-language HS, visual (e.g., a person preparing to shoot) and textual elements (e.g., sighting an illegal immigrant mowing the lawn) jointly convey antagonism. The figure also includes a music parody, integrating derogatory visuals with discriminatory audio lyrics, to showcase contempt for immigrants.

While existing research surveys \cite{rini2020systematic,chhabra2023literature,subramanian2023survey} have largely focused on text-based HS, they often overlook the complexity of multimodal content. Our survey addresses this gap by offering a comprehensive analysis of HS across various digital platforms, including text, visual, auditory, and combined multimodal expressions. We explore the distinct ways HS manifests in these formats, providing insights into their characteristics and moderation challenges. Additionally, we emphasize the critical role of large language models (LLMs) and large multimodal models (LMMs) in moderating HS, given their ability to process and interpret diverse data types. This survey critically evaluates existing solutions, identifies areas for improvement, and advocates for a shift towards multimodal approaches in HS moderation. 

In summary, our paper not only bridges the gap in the existing literature by providing a detailed exploration of multimodal HS but also paves the way for future research in this area. We aim to inspire advancements in HS moderation technology, particularly in the development and refinement of large models, which are imperative for tackling the complex and ever-changing nature of online HS.

\textbf{Paper Collection.} 
We systematically examined research pertaining to the moderation of various types of hate speech, encompassing text, images, videos, and audio. Our search involved keywords such as '\textit{hate speech}', '\textit{multimodal hate speech}', '\textit{hateful memes}', and similar terms, across scholarly platforms like Google Scholar, DBLP, IEEE Xplore, and ACM Digital Library.
Among related research, we further selected state-of-the-art studies, with a particular interest in those using LLMs and LMMs.
Due to the need for a manageable scope, we excluded works that did not leverage LLMs or LMMs, or focused narrowly on regional or multilingual aspects without broader relevance. This decision is not a reflection on the quality or importance of these works but rather a necessity to maintain a focused and coherent survey.



\begin{table*}[t!]
    \scriptsize
    \centering
    \begin{tabular}{lllp{18.5em}lrrr}
        \toprule
         \textbf{Mod.} & \textbf{Dataset} & \textbf{Task} & \textbf{Labels} & \textbf{Source}  & \textbf{\# Records} \\
         \midrule
         \multirow{36}{2em}{Text} & WZ-LS \cite{waseem2016hateful} & Det. & \textbf{[M.C.]} Sexism, Racism, Neither & Twitter & 16,914 \\
         \cmidrule{2-6}
         & GHC \cite{kennedy2018gab} & Det. & \textbf{[M.C.]} VO, HD, CV \textbf{[B]} Implicitness \textbf{[M.C.]} Hate Targets & Forums & 27,665 \\
         \cmidrule{2-6}
         & Stormfront \cite{de-gibert2018hate} & Det. & \textbf{[B]} Hateful & StormFront & 9,916 \\
         \cmidrule{2-6}
         & DT \cite{davidson2017automated} & Det. & \textbf{[M.C.]} Hateful, Offensive, Neither & Twitter & 24,802  \\ 
         \cmidrule{2-6}
         \cmidrule{2-6}
         & Founta \cite{founta2018large} & Det. & \textbf{[M.C.]} Offensive, Abusive, Hateful Speech, Aggressive, Cyberbullying, Spam, Normal & Twitter & 80,000 \\
         \cmidrule{2-6}
         & DynaHate \cite{vidgen2021learning} & Det. & \textbf{[B]} Hateful, \textbf{[M.C.]} Hate Targets, \textbf{[M.C]} Animosity, Derogation, Dehumanization, Threatening, Support & H-M Adv & 41,134 \\
         \cmidrule{2-6}
        & SBIC \cite{sap2019social} & Det. & \textbf{[B]} Offensive \textbf{[M.C]} Hate Targets \textbf{[B]} Intent \textbf{[B]} Lewd \textbf{[B]} Group \textbf{[B]} In-Group & Mixed & 44,671 \\
         &  & Expl. & \textbf{[F.T.]} Implied Statement \\
         \cmidrule{2-6}
         & IHC \cite{elsherief2021latent} & Det. & \textbf{[M.C.]} Implicit, Explicit, Non-Hate \textbf{[M.C.]} Grievance, Incitement, Inferiority, Irony, Stereotypical, Threatening, Others & Twitter & 22,584 \\
         & & Expl. & \textbf{[F.T.]} Implied Statement \\
         \cmidrule{2-6}
         & HateXplain \cite{mathew2021hatexplain} & Det. & \textbf{[M.C]} Hate, Offensive, Normal \textbf{[M.C.]} Hate Targets & Mixed	& 20,148 \\
         & & Expl. & \textbf{[M.L]} Text Rationales/Snippets \\
         \cmidrule{2-6}
         & NACL \cite{masud2022proactively} & Det. & \textbf{[M.C.]} Hate Intensity \textbf{[M.L.]} Hate Spans & Mixed & 4,423 \\
         & & Ctr. & \textbf{[F.T.]} Hate Speech Normalization \\
         \cmidrule{2-6}
         & CONAN \cite{chung2019conan} & Det. & \textbf{[M.C.]} Hate Types \textbf{[M.C.]} Hate Sub-Topic & Synthetic & 14,988 \\
         & & Ctr. & \textbf{[F.T.]} CN Generation \\
         \cmidrule{2-6}
         & Multitarget CONAN \cite{fanton2021human} & Det. & \textbf{[M.C.]} Hate Targets & GPT-2 & 5,003 \\
         & & Ctr. & \textbf{[F.T.]} CN Generation \\
         \cmidrule{2-6}
         & Counter Narratives \cite{das2023hatemm} & Ctr. & \textbf{[F.T.]} CN Generation & YouTube & 9,119 \\
         \midrule
         \multirow{20}{2em}{Img} & MMHS150K \cite{DBLP:conf/wacv/GomezGGK20} & Det. & \textbf{[B]} Hateful & Twitter & 150,000 \\
         \cmidrule{2-6}
         & FHM \cite{DBLP:conf/nips/KielaFMGSRT20} & Det. & \textbf{[B]} Hateful & Synthetic & 10,000 \\
         \cmidrule{2-6}
         & Finegrained FHM \cite{mathias-etal-2021-findings} & Det. & \textbf{[B]} Hateful \textbf{[M.L.M.C]} Protected Category \textbf{[M.L.M.C]} Protected Attacks & Synthetic & 10,000 \\
         \cmidrule{2-6}
         & Misogynous Meme & Det. & \textbf{[B]} Misogynistic \textbf{[B]} Aggressive & Mixed & 800 \\
         & \cite{gasparini2022benchmark} & & \textbf{[B]} Ironic \\
         \cmidrule{2-6}
         & MAMI \cite{DBLP:conf/semeval/FersiniGRSCRLS22} & Det. & \textbf{[B]} Misogyny \textbf{[M.L.M.C.]} Misogynous, Shaming, Stereotype, Objectification, Violence & Mixed & 10,000 \\
         \cmidrule{2-6}
         & UA-RU Conflict \cite{DBLP:conf/case-ws/ThapaSJNR22} & Det. & \textbf{[B]} Hateful & Twitter & 5,680 \\
         \cmidrule{2-6}
         & CrisisHateMM \cite{DBLP:conf/cvpr/BhandariSTNN23} & Det. & \textbf{[B]} Hateful \textbf{[B]} Directed \textbf{[M.C.]} Hate Targets & Mixed & 4,723 \\
         \cmidrule{2-6}
         & RUHate-MM \cite{thapa2024ruhate} & Det. & \textbf{[B]} Hateful \textbf{[M.C]} Hate Targets & Twitter & 20,675 \\
         \cmidrule{2-6}
         & HatReD \cite{DBLP:conf/ijcai/HeeCL23} & Expl. & \textbf{[F.T]} Explanations & Synthetic & 3,228 \\
         \midrule
         \multirow{4}{2em}{Video} & Bangla Hate Videos \cite{junaid2021bangla} & Det. & \textbf{[B]} Hateful & YouTube & 300 \\
         \cmidrule{2-6}
         & HateMM \cite{das2023hatemm} & Det. & \textbf{[B]} Hateful \textbf{[M.C.]} Hate Targets & Mixed & 1,083 \\
         \cmidrule{2-6}
         & MultiHateClip \cite{wang2024multihateclip} & Det. & \textbf{[M.C]} Hateful, Offensive, Normal & YouTube \& Bilibili & 2,000 \\
         \midrule
         \multirow{2}{2em}{Audio} & DeToxy \cite{ghosh2021detoxy} & Det. & \textbf{[B]} Hateful & Mixed & 2M \\
         \cmidrule{2-6}
         \multirow{2}{2em}{} & MuTox \cite{costa2024mutox} & Det. & \textbf{[B]} Hateful & Mixed & 116,000 \\
         \bottomrule
    \end{tabular}
    \caption{Publicly available datasets for HS detection (Det.), HS explanation (Expl.) and counter HS (Ctr.). Abbreviation:  \textbf{M.L.}: multi-label, \textbf{M.C.}: multi-class, \textbf{M.L.M.C.}:  multi-label multi-class, \textbf{B}:  binary, \textbf{F.T}: free-text, \textbf{H-M Adv}: Human-Machine Adversarial. \textit{Note that multilingual HS is out of the scope for the current review}.}
    \label{tab:literature-overview}
\end{table*}

\section{Hate Speech}
\begin{figure}[t!]
\centering
\includegraphics[width=0.90\linewidth]{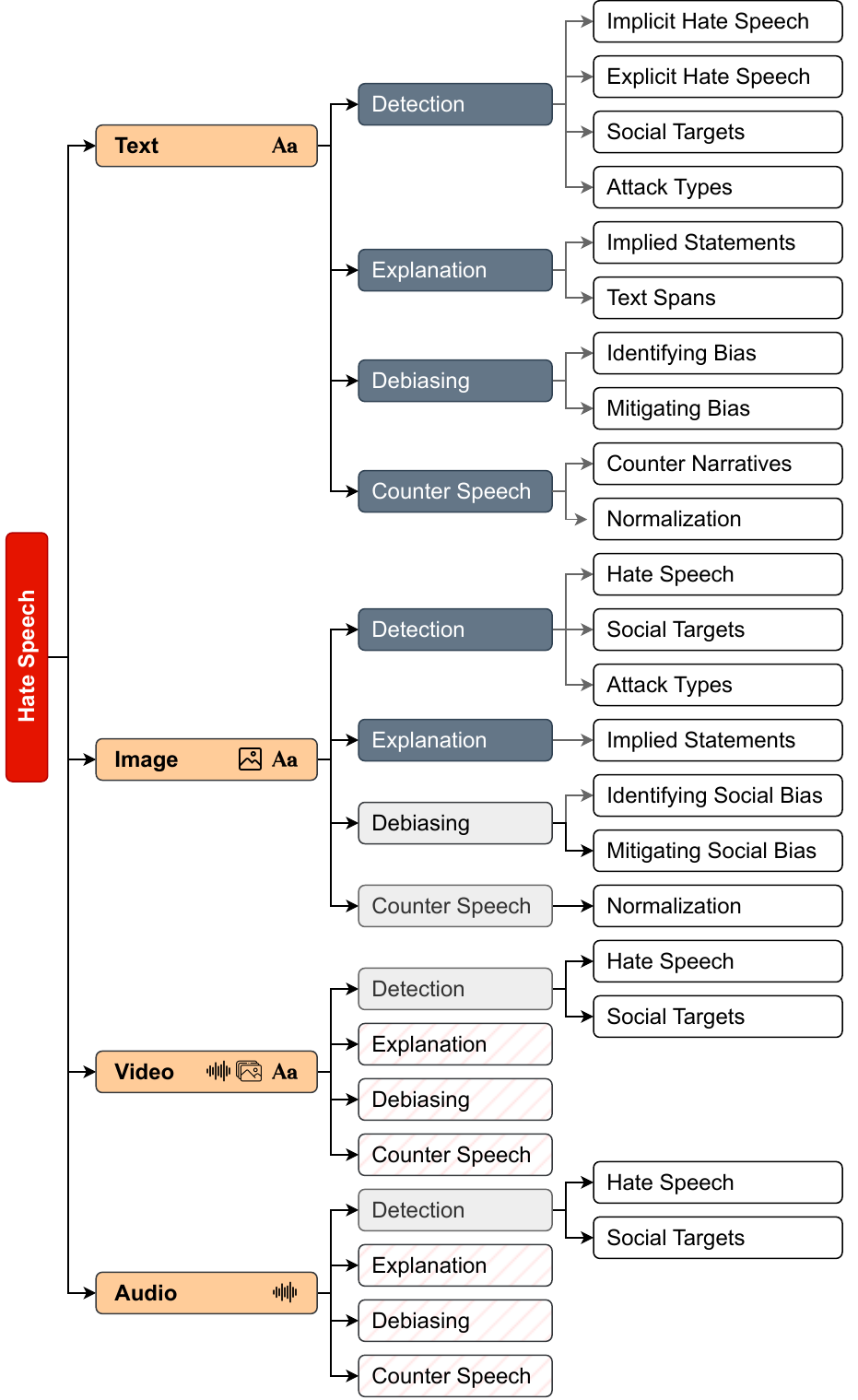}
\caption{Typology of HS based on modalities and tasks. The dark blue boxes are mature areas with multiple studies; light grey boxes are ongoing research, 
and hatched boxes are unexplored topics.}
\label{fig:typology}
\vspace{-3mm}
\end{figure}

HS takes various forms — written text, images, spoken words, and multimedia content — each posing risks of violence, animosity, or prejudice against specific groups. This section reviews existing literature on HS, categorizing it into text-based, image-based, video-based, and audio-based types. 
For each HS form, we provide a detailed categorization across four tasks: \textit{detection}, \textit{explanation}, \textit{debiasing}, and \textit{counter-speech}. Detection identifies hateful content, forming the basis for further actions. Explanation promotes transparency by clarifying why content is flagged, building trust in automated systems. Debiasing is essential to refine detection systems, ensuring fairness and reducing bias. Counter-speech involves taking proactive steps to mitigate the impact of hate speech, fostering healthier online dialogue. Although these tasks address different aspects, they collectively form the foundation of an effective content moderation strategy, highlighting both the interconnectedness of HS forms and the research gaps in advancing multimodal HS moderation.
Figure \ref{fig:typology} illustrates the range of online HS forms. 
Additionally, Table 1 lists publicly accessible HS datasets in different modalities, providing researchers with essential resources for HS moderation.




\subsection{Text-based Hate Speech}
Text-based HS encompasses written or typed expressions manifested across online platforms, such as social media posts \cite{waseem2016hateful,founta2018large}. Recent studies explored diverse aspects of hate and derogatory language, focusing on implicit HS \cite{sap2019social}, targeted groups \cite{kennedy2018gab, yoder2022hate}, and types of attacks \cite{elsherief2021latent}. As HS detection models improve, it becomes imperative to understand and explain their decision-making processes, mitigating unintended bias \cite{garg2023handling}. Additionally, some research shifted towards proactive strategies, including countering HS \cite{masud2022proactively}.



The detection of text-based HS poses numerous challenges. Detecting hate speech (HS) in a single statement often requires understanding dark humor and cultural nuances \cite{hee2024understanding}. HS can express underlying intent through sarcasm, irony, or cultural references, which may not be immediately apparent. Linguistic variations, such as slang, dialects, and unconventional language use, further complicate the task. The challenge intensifies when considering the broader context of an utterance \cite{nagar2023towards,yu2022hate}, as statements that seem neutral in isolation may reveal hateful intent when viewed within a conversation. Conversely, what appears offensive might be harmless in context. Therefore, context-aware models are essential for accurately identifying HS by analyzing both individual statements and their surrounding situational context. Expanding the analysis to converssations, such as Reddit threads or WhatsApp chat, adds additional layers of complexity 
\cite{naseem2019deep}. The intent behind a single message can shift based on prior exchanges and the overall tone of the conversation. Furthermore, user-specific features may be important for HS detection \cite{qian2018leveraging}. Data such as a user's posting history, profile, and behavior provide valuable context for identifying hate speech, though using such data raises ethical concerns, particularly regarding privacy.

\subsection{Image-based Hate Speech}
Image-based HS utilizes visual elements, such as photographs, cartoons, and illustrations, to propagate hate or discrimination against specific groups. A common manifestation of this HS form is memes, which typically consist of images combined with short overlaid text. Although memes often serve humorous or satirical purposes, they are increasingly used to spread hateful content online \cite{DBLP:conf/nips/KielaFMGSRT20}. Recent studies have developed datasets for identifying HS \cite{DBLP:conf/wacv/GomezGGK20}, specific targets \cite{mathias-etal-2021-findings} and types of attacks \cite{DBLP:conf/semeval/FersiniGRSCRLS22} within these memes. Beyond detection, new approaches analyze and mitigate bias in image-based HS detection models \cite{DBLP:conf/ijcai/HeeCL23,DBLP:conf/emnlp/LinLMC23}. Additionally, new methodologies are emerging to counteract HS transmitted through memes \cite{DBLP:journals/corr/abs-2311-06737}.

Image-based HS presents new challenges due to the subtlety of offensive messages concealed within multiple modalities. Images, often embedding symbols, memes, or culturally specific visual cues, require deep cultural and contextual understanding for accurate interpretation. The visual elements and text can subtly imply meanings not immediately evident \cite{DBLP:conf/nips/KielaFMGSRT20}. For example, Figure 1 depicts a man with a gun and text suggesting hostility towards immigrants. Differentiating humour from hate in memes is particularly challenging, influenced by varying cultural, societal, and personal perspectives \cite{schmid2023humorous}. 

\subsection{Video-based Hate Speech}
Video-based HS presents a complex challenge, comprising a blend of visuals, audio tracks, and/or textual elements. This form of HS ranges from professionally produced propaganda to amateur videos on social media platforms like YouTube and TikTok \cite{das2023hatemm,wang2024multihateclip}. The engaging nature of video content and its easy dissemination across digital networks significantly heighten its potential for harm. Echoing the concerns of image-based HS, video-based HS also contributes to the normalization of hateful ideologies and can profoundly influence public opinion. Contemporary research primarily focuses on identifying video-based HS and categorizing its various subtypes \cite{wu2020detection}. Nonetheless, the amount of research on video-based HS is less developed than text-based and image-based HS, particularly in areas such as analyzing and mitigating model bias, elucidating decision-making processes, and devising counterstrategies. These gaps, likely stemming from the rapid pace of technological advancements and evolving digital trends, underscore the need for further research to promote a more harmonious online environment.


Detecting hate speech (HS) in videos is challenging and resource-intensive because it requires understanding various elements, including text, images, and audio, both independently and in combination. Each component can independently contain hateful content, further complicating the detection process. The duration of videos further exacerbates this challenge, as longer content necessitates more extensive review and analysis, with potential shifts in context over time. Moreover, subtle visual cues and sophisticated editing techniques can be employed to discreetly embed hate messages, making their detection by automated tools particularly challenging. Additionally, video content analysis requires considerable computational resources and time, posing a substantial challenge for organizations to detect and address HS in video formats.

\subsection{Audio-based Hate Speech}
Audio-based HS entails the analysis of sound waves to discern elements such as pitch, intonation, and the contextual meaning of spoken words. This form of HS can originate from a variety of audio channels, including real-time conversations, podcasts, and other forms of audio media. The methodologies for addressing audio-based HS are diverse, targeting different facets of the issue. For instance, \citet{6376654} employed a straightforward keyword-based approach to identify segments of HS, while \citet{9287133} engaged in a detailed classification of offensive categories in audio-based HS, showcasing a nuanced method of understanding and categorizing this form of HS. This research area is still in its developmental stages, partly due to the scarcity of dataset. Nonetheless, recognizing the variety and significance of the approaches and techniques employed in this field is imperative. This recognition not only sheds light on the current state of research but also illuminates potential avenues for future exploration.


Detecting HS in audio recordings presents unique challenges, primarily related to the transcription and interpretation of spoken words. The accuracy of speech recognition is crucial, especially when dealing with diverse accents, background noise, or poor audio quality. Additionally, the tone and intonation of spoken language play a significant role in conveying intent, which can substantially alter the meaning of words. This aspect poses a challenge for detection based solely on text transcripts, as subtle nuances in vocal expression may be lost during transcription. Moreover, non-verbal audio elements, such as sound cues or background noises, are pivotal in contextualizing speech. However, these elements are often difficult to interpret using automated methods.

\section{Methodology}
\label{sec:method}


This section reviews the state-of-the-art methodologies that have significantly contributed to primary areas of HS research, particularly those involving large models. First, we discuss the recent capabilities of large models (Section \ref{sec:methodology-large-models}). Subsequently, we explore studies in four important HS areas: \textit{detection} (Section \ref{sec:methodology-detection}), \textit{explanation} (Section \ref{sec:hate-speech-explanation}), \textit{debiasing} (Section \ref{sec:methodology-debiasing}), and \textit{counter-speech} (Section \ref{sec:methodology-counter-speech}), focusing on works using large models. This review highlights the emerging trends, providing insights into how large models can be used to understand and address HS in its various forms.

\subsection{Large Models}
\label{sec:methodology-large-models}

The emergence of large foundation models, such as LLMs and LMMs, marks a significant milestone in artificial intelligence research, showcasing unprecedented capabilities in understanding and generating data across different formats  \cite{zhao2023survey}. LLMs are designed to excel in language understanding and text generation  \cite{touvron2023llama}. In contrast, LMMs are adept at processing and interpreting various data types, including visual, textual, and auditory inputs, enabling a broader spectrum of applications \cite{yang2023dawn}. These foundation models have opened new avenues for identifying and mitigating hateful content, which requires nuanced understanding of language and context.

Here, we regard LLMs and LMMs as models with several billion parameters, aligning with the definition widely accepted and analyzed in numerous studies of large-scale models \cite{luo2023zero}.


\subsection{Hate Speech Detection}
\label{sec:methodology-detection}


The leading detection techniques for HS vary according to the modality of the content, encompassing approaches from transformer-based models to spectrogram-based classification models. 
For text-based HS detection, approaches range from embedding-based methods to advanced neural models \cite{cao2020deephate, davidson2017automated, badjatiya2017deep, fortuna2018survey}. AngryBERT \cite{awal2021angrybert} fine-tunes BERT using a multi-task learning strategy for binary text HS detection.
PromptHate \cite{DBLP:conf/emnlp/CaoLC022} combines demonstration sampling and in-context learning to fine-tune RoBERTa for hateful meme detection.
In audio-based HS detection, ensemble techniques such as AdaBoost, Naive Bayes, and Random Forest have been employed. \cite{boishakhi2021multi,ibanez2021audio}.
CNNs are also used to convert audio into spectrograms \cite{Medina2022}, with self-attentive CNNs extracting audio features \cite{9616001}. For video-based HS detection, a combination of BERT, ViT, and MFCC has been used for text, image, and audio modality analysis, respectively \cite{das2023hatemm}.
Note that audio-based and video-based HS detection are emerging areas with significant potential for future advancements.

Transformer-based models have significantly advanced the detection of text-based and image-based HS; yet they encounter challenges. For text-based models, a major hurdle is generalizing to out-of-distribution datasets, often hindered by limited vocabulary and the rarity of implicit HS in many datasets \cite{ocampo2023depth}. To overcome this, recent initiatives include adversarial HS generation and in-context learning with LLMs. \citet{ocampo2023playing} introduced a method using GPT-3 to generate implicit HS, aiming to both challenge and improve HS classifiers. Concurrently, \citet{wang2023large} developed a technique for optimizing example selection for in-context learning in LLMs.

In image-based HS, the primary challenge lies in deciphering implicit hate messages within memes. This often stems from the loss of information during the extraction of text-based features from images, a common step in many methodologies \cite{DBLP:conf/mm/LeeCFJC21,pramanick2021momenta,DBLP:conf/emnlp/CaoLC022}. Furthermore, the implicit HS in memes can be concealed by seemingly unrelated text and images, as illustrated in Figure \ref{fig:hate-speech-modalities}. To address these challenges, recent strategies include employing LMMs with prompting techniques and/or knowledge distillation. Pro-Cap \cite{DBLP:conf/mm/CaoHKCL023} addresses the issue of information loss in image-to-text conversion by prompting an LMM in a QA format, enhancing the generated caption's quality and informativeness. To tackle the problem of disconnected text and images, MR.HARM \cite{DBLP:conf/emnlp/LinLMC23} utilizes an LMM to generate potential rationales. These rationales are subsequently employed to fine-tune supervised HS classification systems through knowledge distillation, improving the detection of hateful memes.

\subsection{Hate Speech Explanation}
\label{sec:hate-speech-explanation}
A major challenge in contemporary HS detection methods is their lack of explainability in decision-making processes. Explainability is crucial for fostering user trust and facilitating systems that require human interaction \cite{balkir2022necessity}. One proposed solution involves training supervised models that not only categorize HS but also provide rationales for these classifications. \citet{sap2019social} and \citet{elsherief2021latent}  developed text-based HS datasets with human-annotated explanations, setting benchmarks for identifying underlying hate. Similarly, \citet{DBLP:conf/ijcai/HeeCL23}  compiled a dataset for hateful memes, complete with human-annotated explanations and benchmarks. However, collecting human-written explanations is not only time-consuming but also susceptible to individual biases. Moreover, it involves the risk of subjecting human annotators to prolonged exposure to HS, which can have adverse psychological effects.

Recent studies have delved into employing LLMs to generate plausible and meaningful explanations for HS. For instance, \citet{DBLP:conf/ijcai/WangHACL23} demonstrated that GPT-3 can craft convincing and effective explanations for HS, a finding substantiated by extensive human evaluations. Additionally, {HARE} \cite{yang2023hare} introduces two prompting methods that generate rationales for HS, enhancing the training of HS detection models and improving its performance. This approach presents an alternative means of developing insightful explanations, while simultaneously mitigating the risks associated with prolonged human exposure to HS. Nevertheless, this area of research is still nascent, thus presenting numerous opportunities for further investigation and development.


\subsection{Hate Speech Debiasing}
\label{sec:methodology-debiasing}

Bias in HS detection models poses a significant risk to their effectiveness and fairness, leading to potential adverse impacts on individuals and society. Addressing this, numerous studies have focused on identifying and mitigating bias in these models. \citet{sap2019risk} found that two widely-used corpora exhibit bias against African American English, which increases the likelihood of classifying tweets in this dialect as hateful. 
\citet{DBLP:conf/www/HeeLC22} conducted a quantitative analysis of modality bias in hateful meme detection, observing that the image modality significantly influences model predictions. Their study also highlighted the tendency of these models to generate false positives when encountering specific group identifier terms.


Beyond merely identifying biases, various studies have introduced innovative methods to reduce these biases within models. \citet{kennedy2020contextualizing} developed a regularization technique utilizing SOC post-hoc explanations to address group identifier bias. Similarly, \citet{DBLP:journals/ipm/RizziGSRF23} observed that models exhibit biases towards terms linked with stereotypical notions about women, such as \textit{dishwasher} and \textit{broom}. To counteract this, the authors proposed a bias mitigation strategy using Bayesian Optimization, which effectively lessened the bias while preserving overall model performance.

These efforts underscore the critical importance of not only recognizing, but also actively mitigating bias. This is especially vital as large models increasingly dominate the landscape for generating explanations and enabling transfer learning. 




\subsection{Counter Speech}
\label{sec:methodology-counter-speech}


The approach to countering HS focuses on generating non-aggressive responses that either reduce the spread of HS or transform it into respectful and inoffensive speech. Recent research categorizes counter-speech into various response types and emphasizes the importance of contextual understanding. \citet{yu2023fine} developed a taxonomy of responses to HS, showcasing the diversity of counter-speech tactics. \citet{mathew2019thou} proposed context-specific strategies such as narrative persuasion and active rebuttal. CONAN \cite{chung2021towards} focused on generating counter-narratives that challenge hate directed at marginalized groups using reliable evidence, logical arguments, and diverse perspectives. These non-aggressive strategies reduce the spread of hate speech and foster positive discourse. Beyond generating non-aggressive responses, other approaches involve diminishing (i.e., normalization) or eliminating (i.e., correction) the level of hate in HS. NACL \cite{masud2022proactively} used neural networks to paraphrase hate speech, effectively lowering the intensity of hate. \citet{DBLP:journals/corr/abs-2311-06737} prompted LMMs to correct HS in memes by replacing hateful text with positive and respectful language.

These studies underscore the critical role of generative models in annotating and developing counter-speech strategies. This further signifies the future opportunities of LLMs and LMMs in enhancing approaches to combat hate speech.

\section{Challenges}

In the dynamic realm of research, especially in areas related to user-generated content and online harmfulness, numerous challenges persist that shape the trajectory and emphasis of scholarly investigations. These challenges, ranging from technical to ethical, define the landscape in which research on HS moderation and detection operates.


\paragraph{Data Complexity, Quality, and Sourcing.} 
The subtlety of some hate speech, known as implicit hate speech, presents a considerable challenge in identifying and understanding the underlying intent, as these intents can hide within seemingly neutral language or actions~\cite{sap2019social,ocampo2023depth,DBLP:conf/nips/KielaFMGSRT20}. This difficulty highlights the complexity of human communication and biases, where hateful messages can be conveyed indirectly or through coded language.
Furthermore, sourcing data from diverse platforms such as Gab, YouTube, and 4chan introduces difficulties in standardization and interpretation \cite{mariconti2019you}. Additionally, the uneven distribution of hate instances across datasets poses significant obstacles for accurate model training. \cite{cao2020hategan}. These challenges underscore the need for advanced methods capable of navigating the intricate and multifaceted nature of data.


\paragraph{Model Performance and Generalizability.} Recent research highlights the importance of enhancing HS detection models for adaptability in various scenarios and contexts. An exemplary example is making HS detection generalizable and effective across domains \cite{awal2021angrybert}, underscoring the need for models to be versatile and not overly reliant on specific content cues such as domain, region, demography, and more. The development of systems like VulnerCheck \cite{mariconti2019you} exemplifies the demand for models that perform well regardless of the context, and that can adapt to the ever-evolving nature of online material. Such adaptability is crucial for identifying and managing new hateful content, especially such designed to bypass advanced AI technologies. The adoption of technologies, like Few-Shot Learner (FSL), for quick adaptation to this evolving landscape is a promising direction.\footnote{\url{https://ai.meta.com/blog/harmful-content-can-evolve-quickly-our-new-ai-system-adapts-to-tackle-it/}} However, it is imperative that these technologies not only understand the content, but also integrate critical aspects of cultural, behavioral, and conversational contexts.

\paragraph{Expression and Modality Variabilities.} 


Research has highlighted the complexities of interpreting hate speech (HS) across various modalities \cite{boishakhi2021multi, DBLP:conf/nips/KielaFMGSRT20}. In text-based HS, implicit hate messages often use dark humor or sarcasm to obscure their true intent, making detection particularly challenging \cite{sap2019social,elsherief2021latent}. For image-based HS, models face difficulty identifying subtle cross-modality nuances that convey the underlying message \cite{DBLP:conf/www/HeeLC22, DBLP:journals/ipm/RizziGSRF23}. These challenges become even more pronounced with audio and video content due to factors such as accents, background noise, inconsistent audio quality, and the inherently ambiguous nature of toxic content \cite{9616001}. Additionally, the potential for misinterpretation and the models' sensitivity to specific trigger words exacerbate these issues \cite{sridhar2022explaining}. Addressing these variabilities is essential and calls for dedicated efforts in future research.

\paragraph{Contextualization.} 
Effective hate speech moderation requires a nuanced understanding of the HS context. Multi-turn interactions in social media conversations, like Reddit threads or Twitter discussions, play a key role in detecting implicit hate speech \cite{ghosh-etal-2023-cosyn} and generating counter speech \cite{yu-etal-2022-hate}. Additionally, Meng et al. proposed DRAG++, a model that predicts hate intensity by analyzing both the content and the full context of conversation threads \cite{meng2023predicting}. Furthermore, geographic-specific factors, such as local slang and cultural differences, influence a model’s ability to generalize across different regions \cite{lee-etal-2023-hate}.
These challenges highlight the need for sophisticated algorithms capable of interpreting language within its contextual usage, thereby enhancing the accuracy and effectiveness of hate speech moderation strategies.


\paragraph{Emerging Domains.} Exploring new and evolving fields, such as the metaverse, presents a distinct set of challenges \cite{Medina2022}. The core of these challenges lies the need to adapt current HS detection methods to new contexts and to develop new strategies specifically designed for the unique characteristics of these platforms. The dynamic and immersive nature of these emerging environments necessitates a re-evaluation and potential re-engineering of current HS detection and mitigation strategies. Future research requires a deep understanding of both technological advancements and the social dynamics within these virtual spaces to ensure effectiveness in detecting and mitigating HS within these evolving digital landscapes.



\paragraph{Bias and Ethical Concerns.} Addressing bias and upholding ethical considerations in HS detection systems poses a significant challenge. Several studies have highlighted these concerns, introducing functional tests for evaluating HS detection models \cite{rottger2022multilingual, ng2024sghatecheck, rottger2020hatecheck}. These challenges are not purely technical but also moral, underscoring the importance of ensuring HS systems operate equitably and do not perpetuate societal biases. Developing responsible HS systems, therefore, require a multidisciplinary approach that combines technical expertise with ethical and societal awareness, ensuring alignment with ethical standards and societal values.


In summary, the areas of HS detection and moderation are confronted with multifaceted challenges. These arise from the inherent complexities of data, technological limitations, modality variabilities, dataset biases, and the uncharted territories of emerging domains like the metaverse. To effectively navigate these obstacles, a concerted and multidisciplinary effort is essential. It calls for the development of methodologies that are not only sophisticated and robust but also highly adaptable. Such methodologies must be capable of contending with the dynamic and often unpredictable nature of user-generated content and online interactions. The future of this field hinges on our ability to continuously evolve and innovate, ensuring that our approaches remain relevant, effective, and ethically sound in an ever-changing digital landscape.


\section{Future Directions}

\paragraph{Cross-Modality Context Understanding.} As hate speech extends beyond mere text to encompass multiple forms of media (multimodality), it becomes crucial for models to have a proficient understanding of context across modalities. Hence, it is imperative that models not only identify hateful content within text or images separately, but also grasp how the combination of text and images can alter the message \cite{DBLP:conf/nips/KielaFMGSRT20}. For instance, an image that is benign on its own might become hateful when paired with specific text. Research could focus on developing models that more effectively understand context across modalities.

\paragraph{Low-Resource Hate Speech Adaptation.} Domain adaptation between related tasks has gained significant attention.
In the domain of hate speech, an exemplary application is the cross-lingual transfer learning for detecting hate speech across different languages. Winata et al. \cite{winata2022cross} use few-shot in-context learning and fine-tuning techniques to adapt insights from languages with abundant resources to those with fewer resources. Given the widespread presence of hate speech and its relatively consistent definitions across different forms, there is potential to extend knowledge from text-based hate speech with abundant resources to other low-resource forms of hate speech. Future research should aim to develop models capable of pre-training on a broad spectrum of multimodal data, including text, images, audio, and videos, to enhance transfer learning capabilities.

\paragraph{Humour \& Sarcasm Understanding.} Comprehending humor and sarcasm involves recognizing subtle linguistic signals and understanding the broader context, which includes cultural, social, and environmental factors. LLMs are adept at processing language but might not entirely capture these intricate details or fully understand the specific circumstances surrounding a statement. Additionally, humor and sarcasm often hinge on wordplay, double meanings, or ambiguous interpretations. Although LLMs can identify language patterns, they might struggle to differentiate between straightforward and figurative speech. Research efforts can focus on enhancing the capability of LLMs and LMMs to interpret sarcasm and humor, particularly dark humor, which conceals itself within the context of a sentence.

\paragraph{Multicultural Moderation.} A challenge in hate speech detection lies in the varying cultural and contextual cues across different countries and regions. These subtle cues often require a nuanced understanding of local languages, dialects, slang, and social norms. This complexity makes it difficult for automated systems to identify and differentiate hate speech from non-offensive content. \citet{nguyen2023extracting} demonstrated how providing cultural common-sense knowledge can alter GPT-3’s behaviour, leading it to produce more accurate and culturally sensitive questions. Similarly, future research could aim to curate HS dataset with regional culture information and build culturally-aware LLMs and LMMs by injecting and fine-tuning models with cultural knowledge.

\paragraph{Real-Time Monitoring.} The vocabulary of hate speech is constantly changing and evolving, particularly in online spaces. Although adapting to different domains can enhance a model's capacity to apply its knowledge across a range of current datasets, the ongoing development of new slurs, coded terms, and symbolic expressions presents a considerable obstacle to the successful detection of hate speech. Research efforts can focus on continual learning methods that enable these models to be updated regularly while minimising the adjustments to their parameters.

\paragraph{Factual Grounding.} Although current methods in generating HS explanation using large-scale models (refer to Section \ref{sec:hate-speech-explanation}) have shown promise, they still face significant challenges. These large models are prone to ``hallucinations'' producing responses that can be factually incorrect, illogical, or unrelated to the initial prompt \cite{ji2023survey}. Consequently, while these recent advancements are promising, the explanations generated by these models are susceptible to misinformation and require verification. Future research should aim to improve the accuracy and relevance of these explanations, which might involve anchoring the explanations in verifiable facts and developing techniques to identify and rectify any discrepancies.

\section{Conclusion}
We highlighted the advancements in HS moderation, underscoring the pivotal role of LLMs and LMMs. Despite these strides, challenges remain, particularly in inclusivity and nuanced detection. Future research should focus on developing AI methodologies that are more context-aware and ethically governed. This endeavor is not only a technological challenge, but also a moral imperative, necessitating interdisciplinary collaboration. As we advance, it is crucial to ensure that technological advancements are matched with a commitment to responsibility, striving for a digital environment that is secure and welcoming for everyone.

\section*{Acknowledgement}
This research is supported by the Ministry of Education, Singapore, under its Academic Research Fund Tier 2 (Award ID: MOE-T2EP20222-0010). Any opinions, findings and conclusions or recommendations expressed in this material are those of the authors and do not reflect
the views of the Ministry of Education, Singapore.
Tanmoy Chakraborty acknowledges Anusandhan National Research Foundation (CRG/2023/001351) for the financial support. 

\section*{Limitations}

There are two limitations to the scope and coverage of our survey paper.

\textbf{Scope.} In this survey paper, we specifically focus on the role of large models in multimodal ``hateful speech moderation''. We recognize that there is extensive research on toxic and harmful content that is closely related to hatefulness. However, hate speech has a distinct definition that attacks need to discriminate against a group of people based on specific traits such as race, gender and sexual orientation. Hence, while these closely related areas are significant, they fall outside the scope of this survey paper. We also recognize different forms of hate speech can exist in multiple languages, resulting in exciting research on multilingual hate speech. However, the primary goal of this paper is to highlight the evolution and adaptation of hate speech in various forms of digital content. Therefore, the topic of multilingualism is beyond the scope of this paper.

\textbf{Research Paper Coverage.} Although numerous research studies on hate speech have been conducted over the past decades, we have focused on state-of-the-art works that either employed large models in their studies or pioneered specific hate speech tasks. This selection enables us to maintain the brevity of this survey paper while focusing our discussion on the promising areas of using large models for hate speech moderation.

\bibliography{custom}

\appendix

\end{document}